\documentclass{article}
\usepackage{graphicx}
\usepackage{caption}
\usepackage{subcaption}
\usepackage{amsmath}
\usepackage{amssymb}
\usepackage{xurl}
\usepackage{booktabs}
\usepackage{multirow}
\usepackage{caption}
\usepackage{subcaption}
\usepackage{tablefootnote}
\usepackage{threeparttable}
\usepackage[hidelinks]{hyperref}

     \usepackage[final,nonatbib]{tackling_climate_workshop_style}

\usepackage[utf8]{inputenc} 
\usepackage[T1]{fontenc}    
\usepackage{hyperref}       
\usepackage{url}            
\usepackage{booktabs}       
\usepackage{amsfonts}       
\usepackage{nicefrac}       
\usepackage{microtype}      

\title{Mapping Housing Stock Characteristics from Drone Images for Climate Resilience in the Caribbean} 

\author{Isabelle Tingzon, Nuala Margaret Cowan, and Pierre Chrzanowski\\
The World Bank Group, GFDRR\\
{\tt\small \{tisabelle, ncowan, pchrzanowski\}@worldbank.org}
}

\begin{document}

\maketitle

\begin{abstract}

Comprehensive information on housing stock is crucial for climate adaptation initiatives aiming to reduce the adverse impacts of climate-extreme hazards in high-risk regions like the Caribbean. In this study, we propose a workflow for rapidly generating critical baseline housing stock data using very high-resolution drone images and deep learning techniques. Specifically, our work leverages the Segment Anything Model and convolutional neural networks for the automated generation of building footprints and roof classification maps. By strengthening local capacity within government agencies to leverage AI and Earth Observation-based solutions, this work seeks to improve the climate resilience of the housing sector in small island developing states in the Caribbean.
\end{abstract}

\section{Introduction}
The Caribbean is among the world's most climate-vulnerable regions due to the prevalence and intensity of extreme climate hazards such as storms, floods, and landslides. Category 5 hurricanes like Dorian, Irma, and Maria have devastated many small island developing states (SIDS) in recent years, leaving widespread trails of loss and destruction across the region. SIDS bear substantial economic costs from climate-extreme events, with the highest degree of damages often sustained in the housing sector \cite{government2020dominica}. Hurricane Maria, for example, destroyed over 90\% of Dominica's housing stock, accumulating costs over 200\% of the nation's GDP \cite{GoCD2017}. As global temperatures continue to rise, extreme weather events will only grow in severity, putting many more vulnerable homes and shelters at risk. 

In response to these challenges, governments and international organizations have developed ambitious climate resilience programs to reduce the adverse effects of extreme climatic hazards in the housing sector \cite{government2020dominica,WorldBank2022}. Climate resilience initiatives generally require comprehensive housing stock information to inform better retrofitting, reconstruction, and relocation plans. However, the traditional house-to-house approach to identifying high-risk buildings can be extremely expensive and time-consuming to implement, prompting the need for more timely and cost-efficient alternatives. 

Recent years have seen a growing interest in using drones for disaster risk reduction and recovery \cite{WorldBank2022,triveno2019coupling}. Previous works have successfully applied deep learning (DL) techniques to very high-resolution (VHR) aerial imagery to extract baseline housing information such as building footprints and rooftop attributes \cite{tingzon2023fusing,buyukdemircioglu2021deep,partovi2017roof,huang2022urban,solovyev2020roof}. However, despite the evident advantages DL and Earth observation (EO)-based technologies, the widespread adoption of these solutions is often hindered by gaps in local capacity to develop and maintain systems for generating baseline exposure datasets. 

This work aims to bridge these gaps by providing government agencies with an end-to-end workflow for rapidly generating critical baseline housing information using VHR drone images. Specifically, we leverage the Segment Anything Model (SAM) for building footprint delineation and convolutional neural networks (CNNs) for roof type and roof material classification. We also evaluate the cross-country generalizability of roof classification models across SIDS to determine the extent to which models trained in one country can be adapted to another. This work is developed under the Digital Earth Project for Resilient Housing and Infrastructure in partnership with the Government of the Commonwealth of Dominica (GoCD) and the Government of Saint Lucia (GoSL) to enhance the climate resilience of the housing sector of SIDS in the Caribbean.

\begin{figure}[t]
    \centering
    \hspace*{-0.5cm}
    \makebox[\textwidth]{
    \begin{subfigure}{.2\textwidth}
      \centering
      \includegraphics[width=1.0\linewidth]{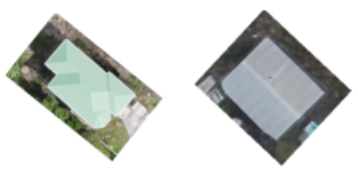}
      \caption{Healthy Metal}
    \end{subfigure}%
    \hspace{0.5cm}
    \begin{subfigure}{.2\textwidth}
      \centering
      \includegraphics[width=1.0\linewidth]{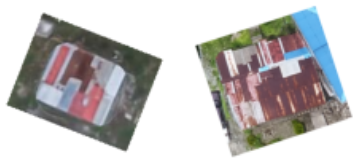}
      \caption{Irregular Metal}
    \end{subfigure}%
    \hspace{0.5cm}
    \begin{subfigure}{.2\textwidth}
      \centering
      \includegraphics[width=1.0\linewidth]{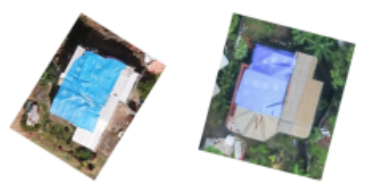}
      \caption{Blue Tarpaulin}
    \end{subfigure}%
    \hspace{0.5cm}
    \begin{subfigure}{.2\textwidth}
      \centering
      \includegraphics[width=1.0\linewidth]{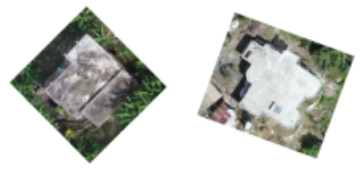}
      \caption{Concrete/cement}
    \end{subfigure}%
    \hspace{0.5cm}
    \begin{subfigure}{.2\textwidth}
      \centering
      \includegraphics[width=1.0\linewidth]{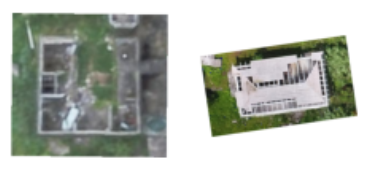}
      \caption{Incomplete}
    \end{subfigure}%
    }
    \\
    \vspace{\floatsep}
    \setcounter{subfigure}{0}    
    \makebox[\textwidth]{
    \begin{subfigure}{.2\textwidth}
      \centering
      \includegraphics[width=1.0\linewidth]{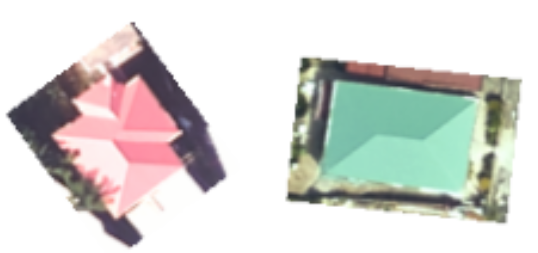}
      \caption{Hip}
    \end{subfigure}%
    \hspace{0.5cm}
    \begin{subfigure}{.2\textwidth}
      \centering
      \includegraphics[width=1.0\linewidth]{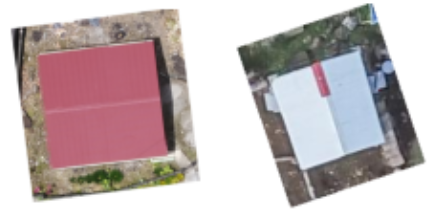}
      \caption{Gable}
    \end{subfigure}%
    \hspace{0.5cm}
    \begin{subfigure}{.2\textwidth}
      \centering
      \includegraphics[width=1.0\linewidth]{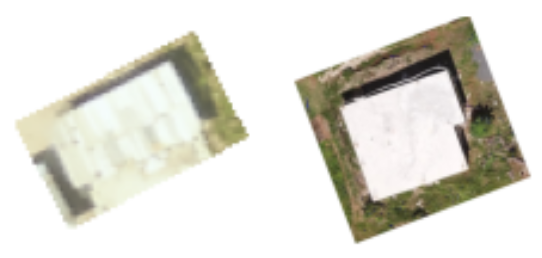}
      \caption{Flat}
    \end{subfigure}%
    \hspace{0.5cm}
    \begin{subfigure}{.2\textwidth}
      \centering
      \includegraphics[width=1.0\linewidth]{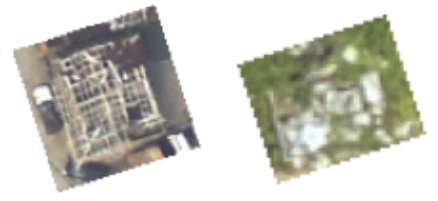}
      \caption{No roof}
    \end{subfigure}%
    }
    \caption{Examples of VHR drone-derived roof image tiles for each of the roof material categories (top row) and roof type categories (bottom row).}
    \label{fig:example-images}
\end{figure}

\section{Data}
To generate our ground truth datasets, we use (1) VHR aerial images in the form of aircraft- and drone-derived optical imagery and (2) building footprints in vector polygons, with Dominica and Saint Lucia as our primary regions of interest. 

\textbf{VHR Aerial Images.} We acquired the following VHR aerial images from partner government agencies GoCD and GoSL, the World Bank Global Program for Resilient Housing (GPRH) \cite{GPRH2019}, and the open data platform OpenAerialMap \cite{smith2015openaerialmap}: (1) aircraft-derived post-disaster orthophotos of Dominica; (2) aircraft-derived orthophotos of Saint Lucia; (3) pre- and post-disaster drone images of 10 villages and cities across Dominica; (4) drone images of 3 districts in Saint Lucia. The spatial resolutions of aircraft-derived images range from 10 to 20 cm/px, whereas the resolutions of drone-derived images range from 2 to 5 cm/px. For more information on the spatial resolution, coverage, and year of acquisition of the aerial images, see Table \ref{tab:aerial-images} in the Appendix. 

\begin{figure}[b]
\centering
\begin{subfigure}{.4\textwidth}
    \centering
    \includegraphics[width=0.9\linewidth]{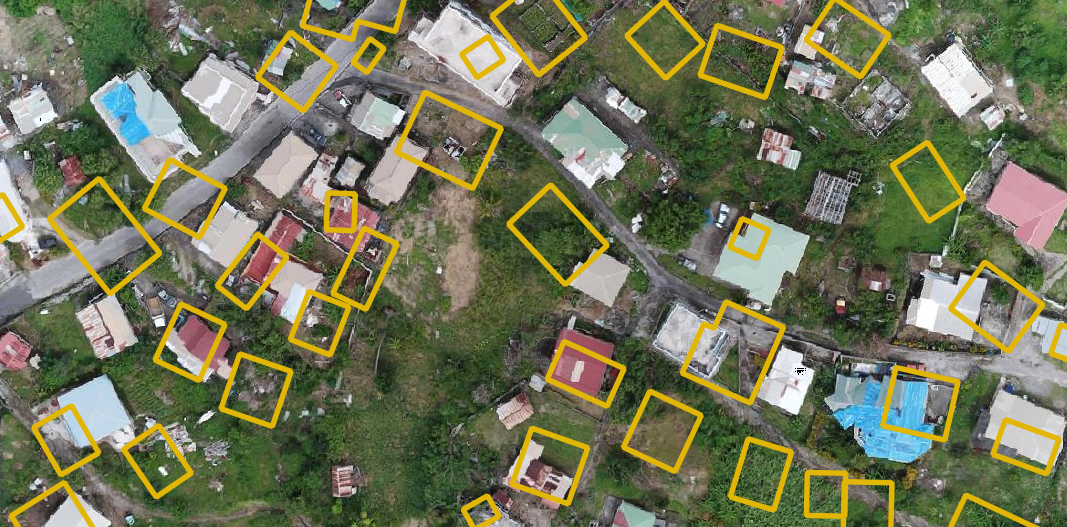}
    \caption{Microsoft}
    \label{fig:enter-label}
\end{subfigure}%
\hspace{0.01\textwidth}
\begin{subfigure}{.4\textwidth}
    \centering
    \includegraphics[width=0.9\linewidth]{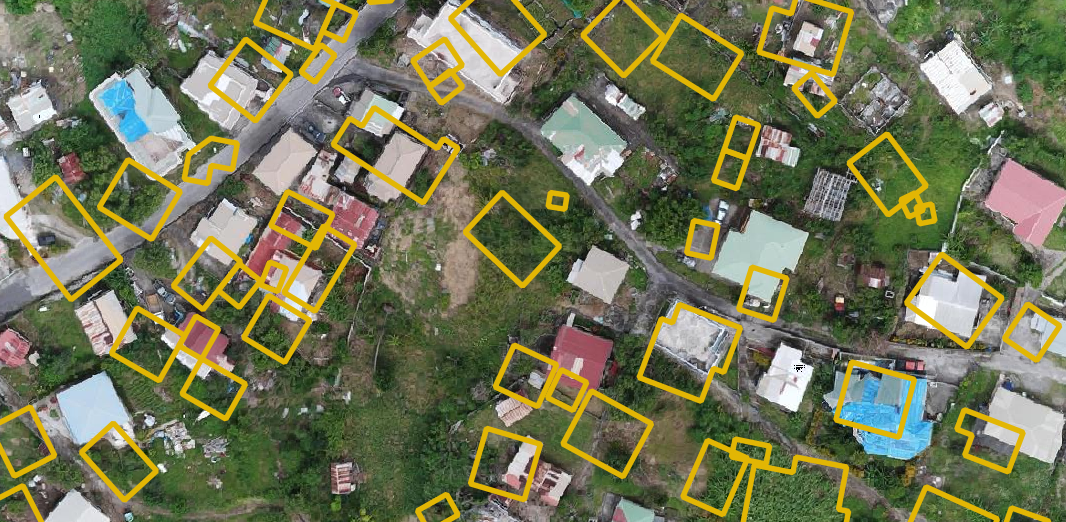}
    \caption{Google}
    \label{fig:enter-label}
\end{subfigure}%
\\
\vspace{0.25cm}
\hspace{0.01\textwidth}
\begin{subfigure}{.4\textwidth}
    \centering
    \includegraphics[width=0.9\linewidth]{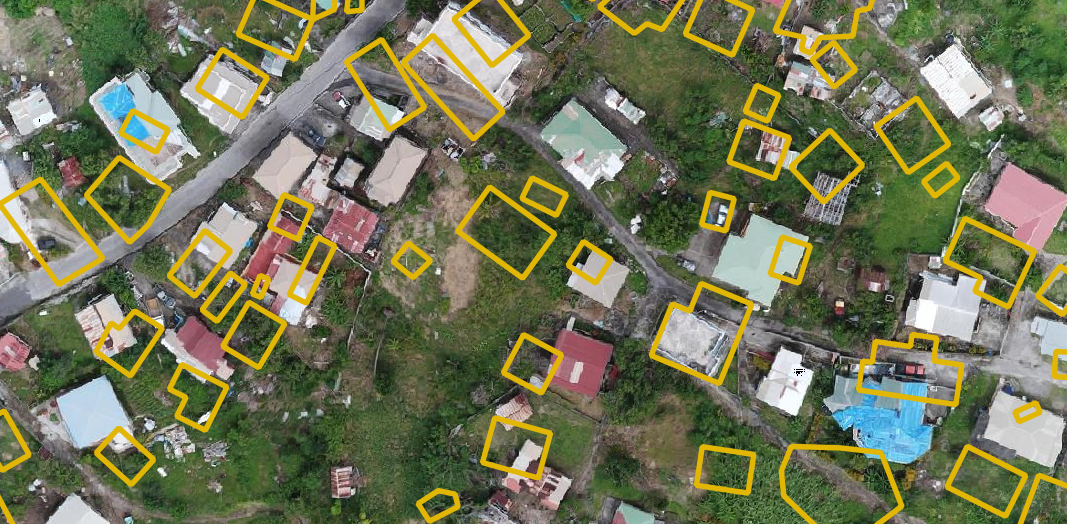}
    \caption{OpenStreetMap}
    \label{fig:enter-label}
\end{subfigure}%
\hspace{0.01\textwidth}
\begin{subfigure}{.4\textwidth}
    \centering
    \includegraphics[width=0.9\linewidth]{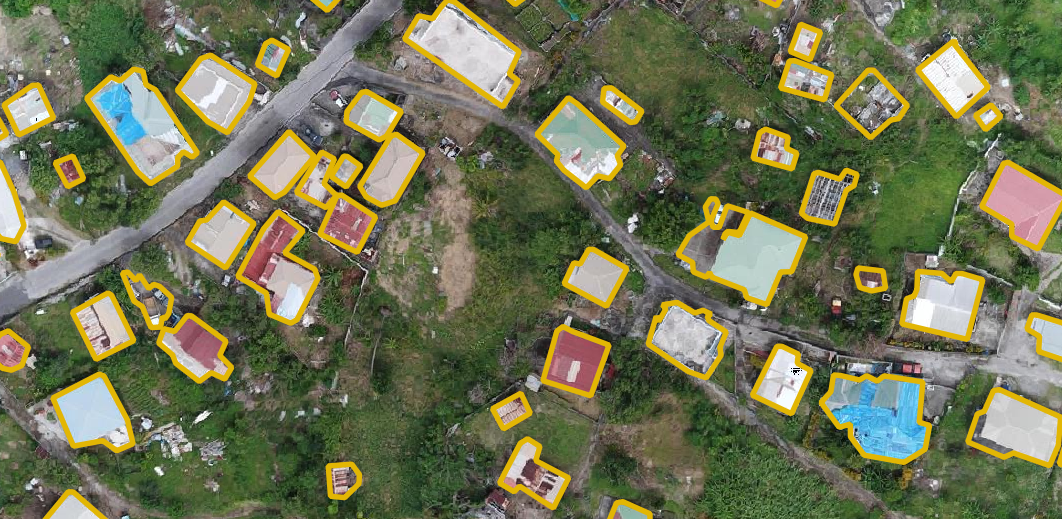}
    \caption{SAM}
    \label{fig:enter-label}
\end{subfigure}%
\caption{Building footprints from (a) Microsoft, (b) Google, (c) OpenStreetMap, and (d) SAM overlaid on a drone image taken in Salisbury, Dominica from OpenAerialMap \cite{smith2015openaerialmap}.}
\label{bldgs}
\end{figure}

\textbf{Building Footprints Data.} We obtained nationwide building footprints delineated from the aircraft-derived orthophotos of Dominica and Saint Lucia from the World Bank. For drone images with no corresponding building footprints, we initially looked to alternative data sources such as OpenStreetMap \cite{osm2023}, Microsoft \cite{mbf2023}, and Google \cite{google2023}. However, we found these publicly available building footprint datasets to be nonviable due to significant misalignment with the underlying drone images, as illustrated in Figure \ref{bldgs}. To address this challenge, we used the Segment Anything Model (SAM) to delineate building instances directly from drone images \cite{kirillov2023segment}. Additional information on the SAM configuration used in this study is detailed in Section \ref{sec:methods}.

\textbf{Rooftop Image Tiles.} For roof classification, we selected approximately 15,000 building footprints across Dominica and Saint Lucia. For each building footprint, we extract the minimal bounding rectangle of the building polygon, scaled by a factor of 1.5, from the corresponding aerial image. We then proceeded to annotate a total of 9,150 buildings in Dominica and 5,175 buildings in Saint Lucia via visual interpretation of VHR aerial images. The buildings are classified based on two attributes: roof type and roof material, the class distributions of which are presented in Table \ref{tab:class-dist}. We note that Saint Lucia does not contain any samples for the "Blue Tarpaulin" category. Figure \ref{fig:example-images} illustrates examples of drone image tiles for each roof type and roof material category. 

\begin{table}[h]
\centering
\small
\caption{The distribution of roof type and roof material labels across Dominica and Saint Lucia.}
\label{table:classdist}
\begin{tabular}{@{}p{0.02cm}p{2.5cm}rrrrrrr@{}}
\toprule
& & \multicolumn{3}{c}{\textbf{Dominica}} & \multicolumn{3}{c}{\textbf{Saint Lucia}} \\ \hline \\[-1.75ex]
& & \textbf{Train} & \textbf{Test} & \textbf{Total} & \textbf{Train} & \textbf{Test} & \textbf{Total} \\
\midrule 
\multirow{4.25}*{\rotatebox{90}{\textbf{\begin{scriptsize} Roof Type\end{scriptsize}}}}  
& \textbf{Gable}   &     2,669        &  653             &    3,322      & 2,347          & 585 & 2,932      \\
& \textbf{Hip}     &  1,579           &     393          &    1,972      &   1,089         & 271 & 1,360     \\
& \textbf{Flat}    &   1,894          & 475               &     2,369     &   456    & 106 & 562     \\
& \textbf{No Roof} &      1,190       & 297              &    1,487      &   269        &  52    & 321  \\
\midrule
\multirow{5.15}*{\rotatebox{90}{\textbf{\begin{scriptsize}
Roof Material\end{scriptsize}}}}  
& \textbf{Healthy metal}   &      1,934       &         482      &   2,416       &  2,396         & 598   & 2,994   \\
& \textbf{Irregular metal} &     1,733        &     432          &     2,165     &  1,113          & 276   &  1,389 \\
& \textbf{Concrete/cement} &  1,240          &   312            &  1,552       &   328        & 75  & 403     \\
& \textbf{Blue tarpaulin}  & 1,094            &   260            &     1,354     &  0          &   0 & 0   \\
& \textbf{Incomplete}      &   1,331          &    332           &  1,663    &   324        &  65  & 389      \\
\midrule
\multirow{-1.075}*{\rotatebox{90}{\textbf{\begin{scriptsize}
Source\end{scriptsize}}}}  
& \textbf{Aircraft}   & 5,936  &  0 & 5,936 &  2,485 & 0 & 2,485   \\ 
& \textbf{Drone}   & 1,396 & 1,818  & 3,214   & 1,676 & 1,014 & 2,690   \\ 
\midrule
& \textbf{Total}   &    \textbf{7,332}         &     \textbf{1,818}          &    \textbf{9,150}      & \textbf{4,161}  & \textbf{1,014} &  \textbf{5,175}   \\ \bottomrule
\end{tabular}
\label{tab:class-dist}
\end{table}

\section{Methods}
\label{sec:methods}
\begin{figure}[t]
    \centering
    \hspace*{-3cm} 
    \includegraphics[width=1.425\textwidth]{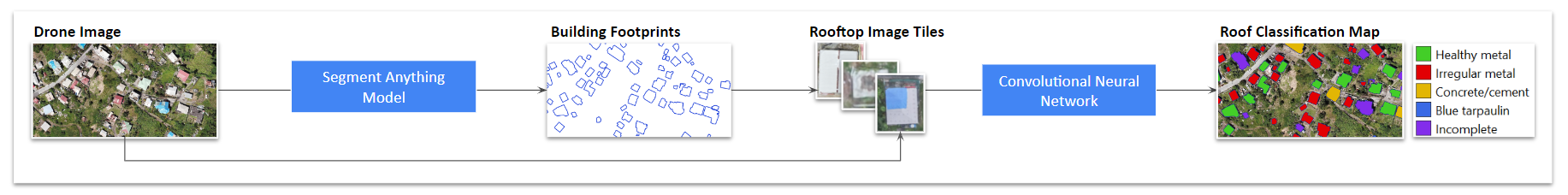}
    \caption{Proposed workflow for the automatic generation of housing stock information from drone images using DL models. }
    \label{fig:workflow}
\end{figure}

This section outlines our workflow for generating critical housing stock information in the Caribbean using drone images, as summarized in Figure \ref{fig:workflow}. 

To extract building footprints from drone images, we use Segment-Geospatial for segmenting raster images based on the Segment Anything Model (SAM) \cite{wu2023samgeo, AIRemoteSensing2023_Zenodo, AIRemoteSensing2023_GitHub, kirillov2023segment}. We also leverage Language Segment Anything \cite{LangSAM2023_GitHub} to combine instance segmentation with text prompts to generate masks of specific objects in the drone images. We set our text prompt to "house", the box threshold (i.e. the threshold value used for object detection in the image) to 0.30, and the text threshold (i.e. the threshold value used to associate the detected objects with the provided text prompt) to 0.30. As a post-processing step, we apply the Douglas-Peucker algorithm to simplify the generated building polygons with a tolerance of $5e^{-6}$ \cite{douglas1973algorithms}. 

Given the VHR aerial images and corresponding building footprint polygons, we proceed with developing our roof classification models. We begin by fine-tuning CNN models pre-trained on the ImageNet dataset \cite{deng2009imagenet} with architectures ResNet50 \cite{he2016deep}, VGG-16 \cite{simonyan2014very}, Inceptionv3 \cite{szegedy2016rethinking}, and EfficientNet-B0 \cite{tan2019efficientnet} using cross-entropy loss for both roof type and roof material classification tasks. The input rooftop image tiles are zero-padded to a square of size 224 x 224 px for ResNet50, EfficientNet-B0, and VGG-16 and 299 x 299 px for InceptionV3. We set the batch size to 32 and the maximum number of epochs to 60, and we use an Adam optimizer with an initial learning rate of $1e^-5$, which decays by a factor of 0.1 after every 7 epochs with no improvement. For data augmentation, we implement horizontal and vertical image flips with a probability of 0.50 and random rotations ranging from $-90^{\circ}$ to $90^{\circ}$. Given that our data is imbalanced (see Table \ref{tab:class-dist}), we implement random oversampling for the minority classes. To prevent overconfident predictions, we apply label smoothing as a regularization technique, with smoothing set to 0.1 \cite{muller2019does}.

\section{Results and Discussion}
For each country in our dataset, we set aside  80\% of the data for training and the remaining 20\% for testing, using stratified random sampling for the test set to preserve the percentage of samples per class as shown in Table \ref{tab:class-dist}. We note that the test sets for both countries are comprised entirely of drone images. Additionally, to test whether geographically diverse training data improves the prediction, we combine the training sets across Dominica and Saint Lucia (henceforth referred to as the "combined" dataset). We report the standard performance metrics F1-score, precision, recall, and accuracy. 

\begin{figure}[b]
    \centering
    \begin{subfigure}{.4\textwidth}
    \centering
    \includegraphics[width=0.9\linewidth]{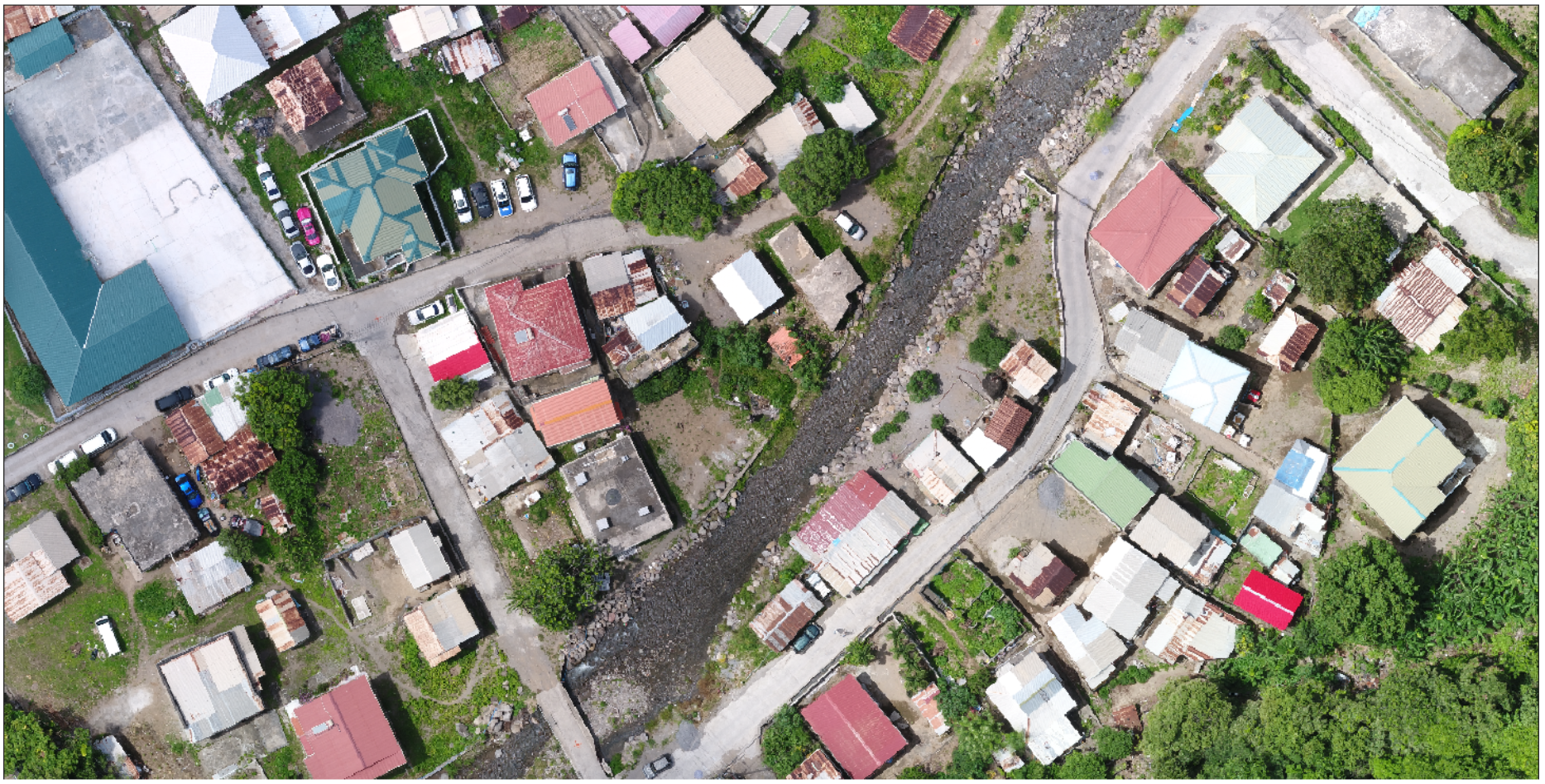}
    \caption{2017 drone image from GoCD}
    \end{subfigure}
    \hspace{0.01\textwidth}
    \begin{subfigure}{.4\textwidth}
    \centering
    \includegraphics[width=0.9\linewidth]{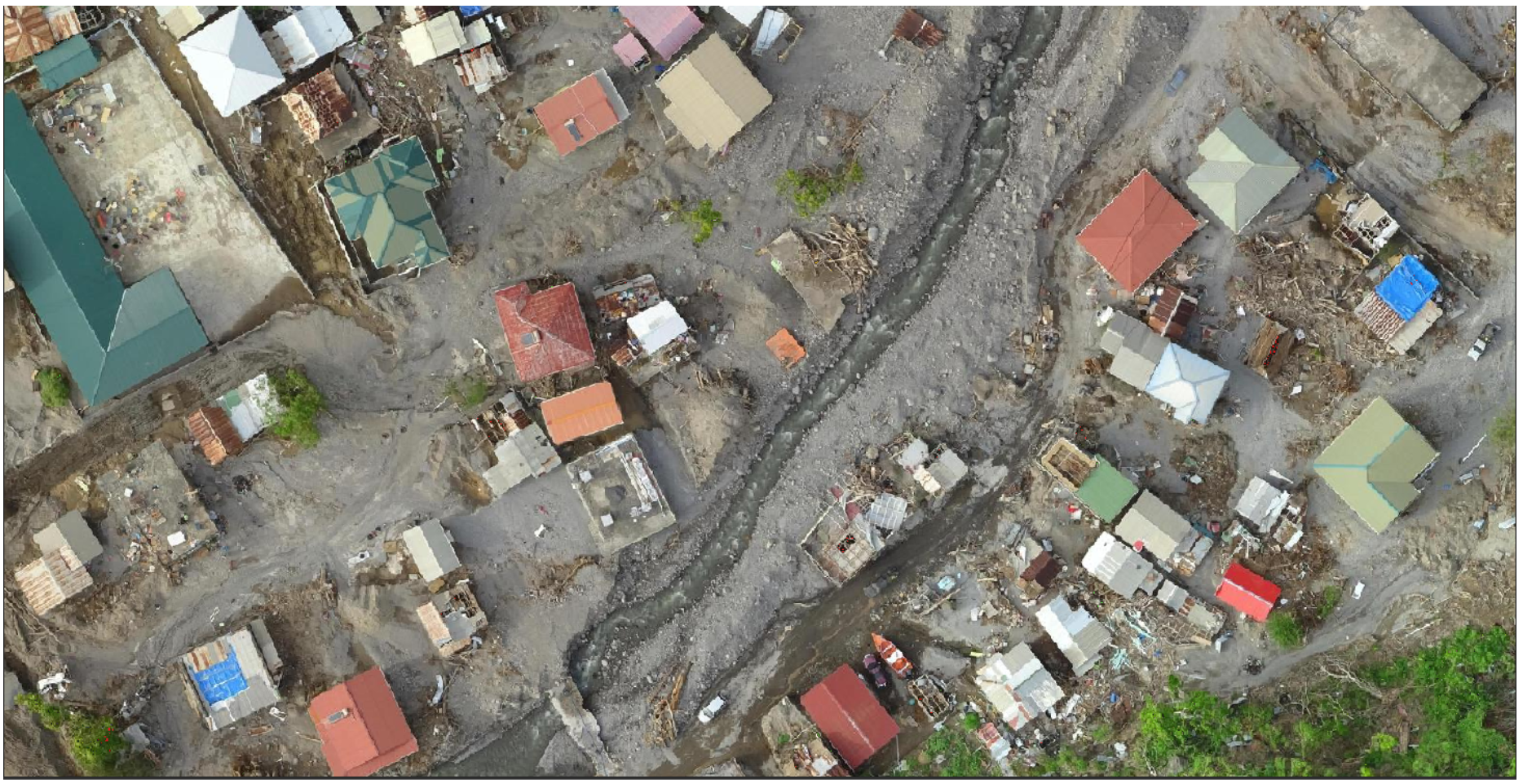}
    \caption{2018 drone image from OpenAerialMap}
    \end{subfigure}
    \\
    \vspace{0.25cm}
    \begin{subfigure}{.4\textwidth}
    \centering
    \includegraphics[width=0.9\linewidth]{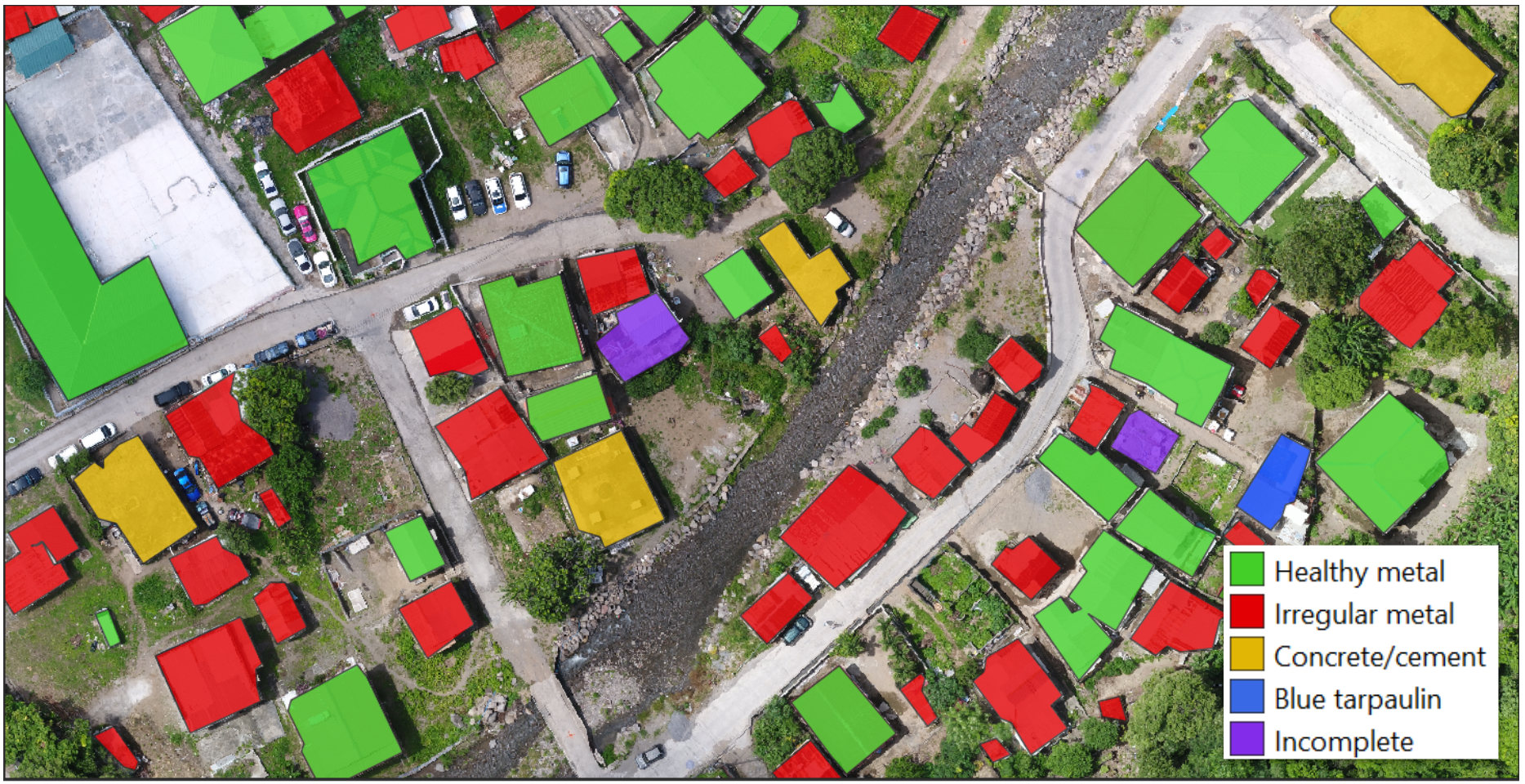}
    \caption{2017 roof material classification map}
    \end{subfigure}
    \hspace{0.01\textwidth}
    \begin{subfigure}{.4\textwidth}
    \centering
    \includegraphics[width=0.9\linewidth]{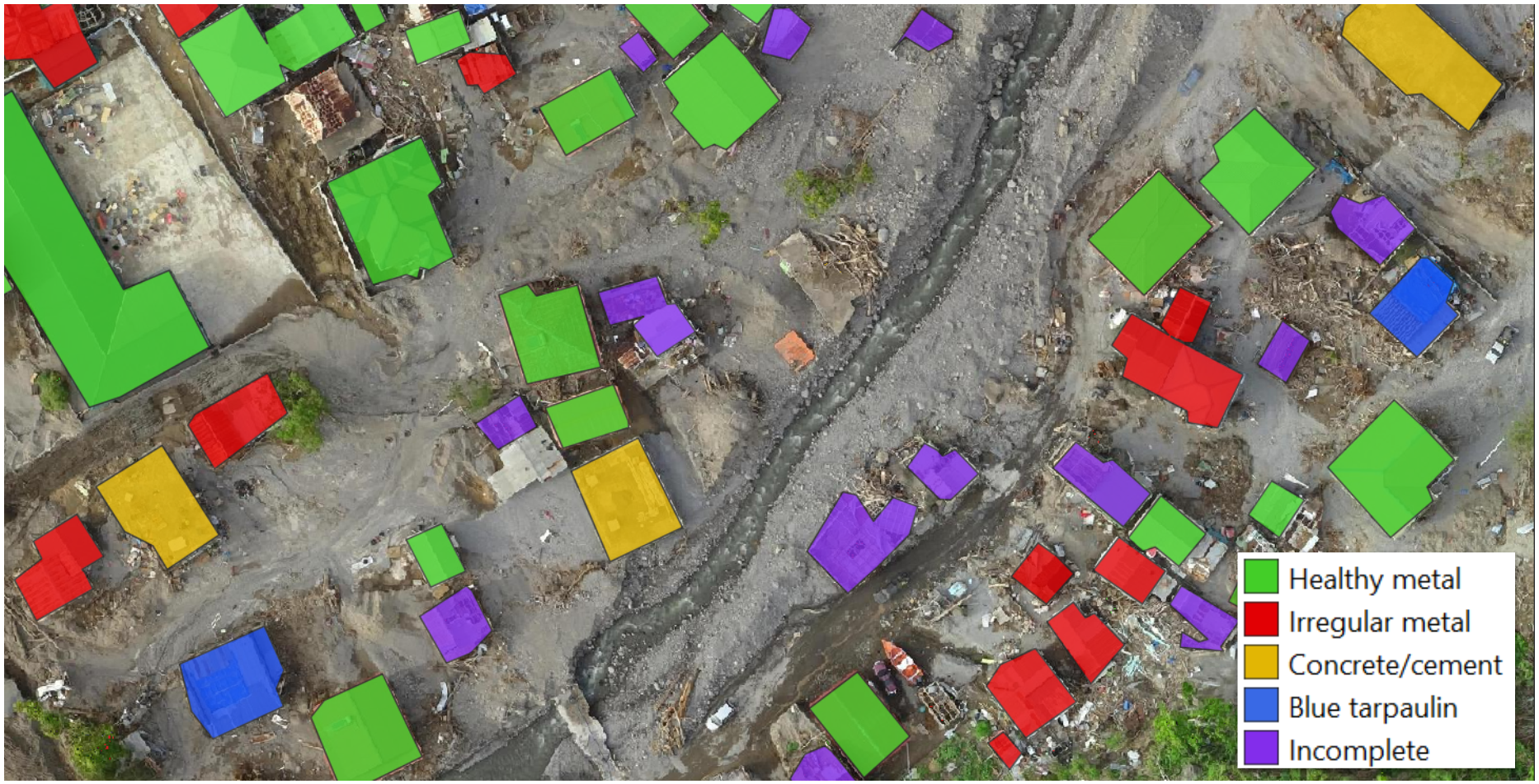}
    \caption{2018 roof material classification map}
    \end{subfigure}
    \caption{Drone images (top) and corresponding roof material classification maps (bottom) of Coulibistrie, Dominica taken before (left) and after (right) Hurricane Maria.}
    \label{fig:example}
\end{figure}

Our results indicate that for Dominica, the best performance is achieved by an EfficientNet-B0 model for roof type classification (F1-score: 87.1\%) and a ResNet50 model for roof material classification (F1-score: 89.5\%). Likewise, for Saint Lucia, the best F1-score  is attained by a ResNet50 model for roof type classification (F1-score: 89.5\%) and an EfficientNet-B0 model for roof material classification (F1-score: 91.71\%). For models trained on the combined dataset (i.e., "combined model"), the best results are obtained by an EfficientNet-B0 model for roof type classification (F1-score: 90.0\%) and an Inceptionv3 model for roof material classification (F1-score: 90.4\%). For the complete results and sample outputs, see Table \ref{table:results1} and Figure \ref{fig:example}, respectively.

Next, we investigate the cross-country generalizability of the best-performing models by evaluating their performance on the designated test sets of each country. As shown in Table \ref{tab:cross-country-results}, our results indicate that for roof type classification, the combined model performs marginally better than models trained using only local data (i.e. data from the same country); however, for roof material classification, we find that locally trained models consistently outperform the combined model. In general, we find that local models demonstrate high levels of variability in performance in out-of-distribution countries, indicating the importance of collecting localized training data and the need for further studies on domain adaptation to reduce performance degradation in the face of geographical distribution shifts.

\begin{table}[h]
\centering
\caption{Cross-country generalization of roof type and roof material classification models.}
\small
\begin{tabular}{llcccc}
\multicolumn{4}{@{}l}{\textbf{(a) Roof Type}}\\
\toprule
\textbf{Training Data} & \textbf{Test Data} & \textbf{F1 score} & \textbf{Precision} & \textbf{Recall} &  \textbf{Accuracy} \\
\midrule
\multirow{2}{*}{Dominica} & Dominica & 87.08 & 87.06 & 87.37 & 86.41  \\
& Saint Lucia & 88.39 & 88.93 & 87.92 & 93.04 \\
 \midrule
 \multirow{2}{*}{Saint Lucia} & Dominica & 82.59 & 84.72 & 81.39 & 82.31  \\
 & Saint Lucia & 89.49 & 94.11 & 85.97 & 94.28\\
 \midrule
 \multirow{2}{*}{Combined} & Dominica & \textbf{88.02} & 88.51 & 88.01 & 87.39  \\
& Saint Lucia & \textbf{91.92} & 93.63 & 90.36 & 95.67  \\
 \bottomrule
\end{tabular}
\medskip \\
\small
\begin{tabular}{llcccc}
\multicolumn{4}{@{}l}{\textbf{(a) Roof Material}}\\
\toprule
\textbf{Training Data} & \textbf{Test Data} & \textbf{F1 score} & \textbf{Precision} & \textbf{Recall} &  \textbf{Accuracy} \\
\midrule
\multirow{2}{*}{Dominica} & Dominica & \textbf{89.50}  & 90.49 & 88.84 & 89.44   \\
& Saint Lucia & 90.83 & 92.04 & 89.93 & 92.94   \\
 \midrule
 \multirow{2}{*}{Saint Lucia} & Dominica & 64.15 & 63.14 & 67.32 & 72.99  \\
 & Saint Lucia & \textbf{91.71} & 93.81 & 90.00  & 93.79   \\
 \midrule
 \multirow{2}{*}{Combined} & Dominica & 88.17 & 89.66 & 87.16 & 87.78   \\
 & Saint Lucia & 90.37 & 93.10 & 88.58 & 93.55  \\
 \bottomrule
\end{tabular}
\label{tab:cross-country-results}
\end{table}

\section{Local Capacity Building}
The overarching goal of this project is to strengthen local capacity in SIDS to leverage AI and EO-based solutions for resilient housing operations. To this end, our team is assisting government agencies in establishing Geographic Information Systems (GIS) units capable of generating, managing, and maintaining large-scale disaster risk datasets. We have partnered with the Humanitarian OpenStreetMap Team (HOT) \cite{team2015hot} to design a training program on how to operate drones, coordinate pilots, process collected imagery, and manage the resulting geospatial datasets. Additionally, we have developed educational resources on geospatial data processing aimed at local government staff, community mappers, and disaster responders, including Google Colaboratory and Jupyter Notebook tutorials demonstrating how to run the SAM and CNN models to quickly generate baseline housing stock data from locally collected drone images \cite{colab_1,colab_2}. 

\section{Conclusion}
This work proposes a workflow for filling baseline exposure data gaps in the Caribbean using VHR drone images and DL. We demonstrate how computer vision and drone-based technologies can be used to rapidly generate housing stock information, including building footprints and roof classification maps, for disaster risk reduction and recovery. Based on our evaluation of the cross-country generalizability of DL models, we urge caution in applying locally-trained models off the shelf to new geographic regions and emphasize the importance of collecting local, highly contextualized training data. We also emphasize the importance of local capacity building, skills development, and co-creation of geospatial datasets in deploying sustainable AI-for-climate solutions, especially in Global South contexts. We conclude this study by urging governments in SIDS to invest in the digital infrastructure and local capabilities needed to sustainably generate and maintain EO-derived housing stock data for climate resilience. 

\section*{Acknowledgments}
This work is developed under Digital Earth for Resilient Housing and Infrastructure in the Caribbean, a World Bank project funded by the Global Facility for Disaster Reduction and Recovery (GFDRR), in partnership with the Government of the Commonwealth of Dominica (GoCD) and the Government of Saint Lucia (GoSL). This study builds on the initial work done by the World Bank's Global Program for Resilient Housing (GPRH). We thank Mike Fedak and Christopher Williams for their assistance in providing data access.

{\small

\begin{thebibliography}{10}

\bibitem{colab_1}
{Building Footprint Delineation for Disaster Risk Reduction and Response (Part
  I)}.
\newblock
  \url{https://colab.research.google.com/github/GFDRR/caribbean-rooftop-classification/blob/master/tutorials/01_building_delineation.ipynb}.

\bibitem{google2023}
{Google Open Buildings}.
\newblock \url{https://sites.research.google/open-buildings/}.
\newblock Accessed on 15.09.2023.

\bibitem{team2015hot}
{Humanitarian OpenStreetMap Team (HOT)}.
\newblock \url{https://www.hotosm.org/}.

\bibitem{mbf2023}
{Microsoft Building Footprints}.
\newblock \url{https://www.microsoft.com/en-us/maps/building-footprints}.
\newblock Accessed on 15.09.2023.

\bibitem{smith2015openaerialmap}
{OpenAerialMap}.
\newblock \url{https://map.openaerialmap.org/}.
\newblock Accessed on 15.09.2023.

\bibitem{osm2023}
{OpenStreetMap Buildings}.
\newblock \url{https://osmbuildings.org/}.
\newblock Accessed on 15.09.2023.

\bibitem{colab_2}
{Rooftop Classification from Drone Imagery for Disaster Risk Reduction and
  Reponse (Part II)}.
\newblock
  \url{https://colab.research.google.com/github/GFDRR/caribbean-rooftop-classification/blob/master/tutorials/02_building_classification.ipynb}.

\bibitem{GPRH2019}
{Global Program for Resilient Housing (GPRH)}.
\newblock
  \url{https://www.worldbank.org/en/topic/disasterriskmanagement/brief/global-program-for-resilient-housing},
  January 2022.

\bibitem{WorldBank2022}
World Bank.
\newblock {Capturing Housing Data in Small Island Developing States}, 2022.
\newblock Washington, DC. Creative Commons Attribution CC BY 4.0.

\bibitem{buyukdemircioglu2021deep}
M~Buyukdemircioglu, R~Can, and S~Kocaman.
\newblock {Deep learning based roof type classification using very high
  resolution aerial imagery}.
\newblock {\em The International Archives of Photogrammetry, Remote Sensing and
  Spatial Information Sciences}, 43:55--60, 2021.

\bibitem{deng2009imagenet}
Jia Deng, Wei Dong, Richard Socher, Li-Jia Li, Kai Li, and Li~Fei-Fei.
\newblock {Imagenet: A large-scale hierarchical image database}.
\newblock In {\em 2009 IEEE conference on computer vision and pattern
  recognition}, pages 248--255. Ieee, 2009.

\bibitem{douglas1973algorithms}
David~H Douglas and Thomas~K Peucker.
\newblock {Algorithms for the reduction of the number of points required to
  represent a digitized line or its caricature}.
\newblock {\em Cartographica: the international journal for geographic
  information and geovisualization}, 10(2):112--122, 1973.

\bibitem{he2016deep}
Kaiming He, Xiangyu Zhang, Shaoqing Ren, and Jian Sun.
\newblock {Deep residual learning for image recognition}.
\newblock In {\em Proceedings of the IEEE conference on computer vision and
  pattern recognition}, pages 770--778, 2016.

\bibitem{huang2022urban}
Xingliang Huang, Libo Ren, Chenglong Liu, Yixuan Wang, Hongfeng Yu, Michael
  Schmitt, Ronny H{\"a}nsch, Xian Sun, Hai Huang, and Helmut Mayer.
\newblock {Urban Building Classification (UBC)-A Dataset for Individual
  Building Detection and Classification From Satellite Imagery}.
\newblock In {\em Proceedings of the IEEE/CVF Conference on Computer Vision and
  Pattern Recognition}, pages 1413--1421, 2022.

\bibitem{kirillov2023segment}
Alexander Kirillov, Eric Mintun, Nikhila Ravi, Hanzi Mao, Chloe Rolland, Laura
  Gustafson, Tete Xiao, Spencer Whitehead, Alexander~C Berg, Wan-Yen Lo, et~al.
\newblock {Segment anything}.
\newblock {\em arXiv preprint arXiv:2304.02643}, 2023.

\bibitem{LangSAM2023_GitHub}
Luca Medeiros.
\newblock {Language Segment-Anything}.
\newblock \url{https://github.com/luca-medeiros/lang-segment-anything}, 2023.

\bibitem{muller2019does}
Rafael M{\"u}ller, Simon Kornblith, and Geoffrey~E Hinton.
\newblock When does label smoothing help?
\newblock {\em {Advances in neural information processing systems}}, 32, 2019.

\bibitem{GoCD2017}
Government of~the Commonwealth~of Dominica.
\newblock {Post-Disaster Needs Assessment Hurricane Maria September 18, 2017}.
\newblock
  \url{https://www.gfdrr.org/sites/default/files/publication/Dominica_mp_012418_web.pdf},
  September 2017.

\bibitem{government2020dominica}
Government of~the Commonwealth~of Dominica.
\newblock {Dominica Climate Resilience and Recovery Plan 2020--2030}.
\newblock
  \url{https://odm.gov.dm/wp-content/uploads/2022/02/CRRP-Final-042020.pdf},
  2020.

\bibitem{AIRemoteSensing2023_Zenodo}
Lucas Osco.
\newblock {AI-RemoteSensing: A collection of Jupyter and Google Colaboratory
  notebooks dedicated to leveraging Artificial Intelligence (AI) in Remote
  Sensing applications}, June 2023.

\bibitem{AIRemoteSensing2023_GitHub}
Lucas~Prado Osco.
\newblock {AI-RemoteSensing}.
\newblock \url{https://github.com/LucasOsco/AI-RemoteSensing}, 2023.

\bibitem{partovi2017roof}
Tahmineh Partovi, Friedrich Fraundorfer, Seyedmajid Azimi, Dimitrios Marmanis,
  and Peter Reinartz.
\newblock {Roof Type Selection based on patch-based classification using deep
  learning for high-resolution Satellite Imagery}.
\newblock {\em International Archives of the Photogrammetry, Remote Sensing and
  Spatial Information Sciences-ISPRS Archives}, 42(W1):653--657, 2017.

\bibitem{simonyan2014very}
Karen Simonyan and Andrew Zisserman.
\newblock {Very deep convolutional networks for large-scale image recognition}.
\newblock {\em arXiv preprint arXiv:1409.1556}, 2014.

\bibitem{solovyev2020roof}
Roman~A Solovyev.
\newblock {Roof material classification from aerial imagery}.
\newblock {\em Optical Memory and Neural Networks}, 29:198--208, 2020.

\bibitem{szegedy2016rethinking}
Christian Szegedy, Vincent Vanhoucke, Sergey Ioffe, Jon Shlens, and Zbigniew
  Wojna.
\newblock {Rethinking the inception architecture for computer vision}.
\newblock In {\em Proceedings of the IEEE conference on computer vision and
  pattern recognition}, pages 2818--2826, 2016.

\bibitem{tan2019efficientnet}
Mingxing Tan and Quoc Le.
\newblock {Efficientnet: Rethinking model scaling for convolutional neural
  networks}.
\newblock In {\em International conference on machine learning}, pages
  6105--6114. PMLR, 2019.

\bibitem{tingzon2023fusing}
Isabelle Tingzon, Nuala~Margaret Cowan, and Pierre Chrzanowski.
\newblock {Fusing VHR Post-disaster Aerial Imagery and LiDAR Data for Roof
  Classification in the Caribbean using CNNs}.
\newblock {\em arXiv preprint arXiv:2307.16177}, 2023.

\bibitem{triveno2019coupling}
Luis Triveno, Sarah Antos, Jan Koers, and Victor Endo.
\newblock {Coupling Imagery from Drones and Street-View with Proper Incentives
  To Promote Sustainable Urban Cadasters in Developing Countries}.
\newblock In {\em Proceedings of the World Bank Conference on Land and Poverty,
  Washington, DC, USA}, pages 25--29, 2019.

\bibitem{wu2023samgeo}
Qiusheng Wu and Lucas~Prado Osco.
\newblock {samgeo: A Python package for segmenting geospatial data with the
  Segment Anything Model (SAM)}.
\newblock {\em Journal of Open Source Software}, 8(89):5663, 2023.

\end{thebibliography}

}

\appendix
\section*{Appendix}
\addcontentsline{toc}{section}{Appendices}
\renewcommand{\thesubsection}{\Alph{subsection}}

\begin{table}[h]
\small
\centering
\caption{Aircraft- and drone-derived aerial images used in this study.}
\begin{tabular}{lrcccr}
\toprule
\textbf{Coverage} & \textbf{Resolution} & \textbf{Year} & \textbf{Source} & \textbf{Data Provider} & \multicolumn{1}{l}{\textbf{\begin{tabular}[c]{@{}r@{}}\textbf{Building} \\ \textbf{Count}\end{tabular}}} \\
\midrule
\begin{tabular}[c]{@{}l@{}}\textbf{Dominica}\vspace{-0.1cm}\\ \begin{scriptsize}(nationwide)\end{scriptsize}\end{tabular} & 20.0 cm/px & 2018-2019 & Aircraft & GoCD & 5,936 \\
\textbf{Colihaut} & 2.7 cm/px & 2017 & Drone & GoCD & 373 \\
\textbf{Coulibistrie} & 2.3 cm/px & 2017 & Drone & GoCD & 158 \\
\textbf{Delices} & 4.3 cm/px & 2018 & Drone & OpenAerialMap \cite{smith2015openaerialmap} & 380 \\
\textbf{Dublanc} & 2.9 cm/px & 2017 & Drone & GoCD & 126 \\
\textbf{Kalinago} & 3.3 cm/px & 2018 & Drone & OpenAerialMap \cite{smith2015openaerialmap} & 102 \\
\textbf{Laplaine} & 4.9 cm/px & 2018 & Drone & OpenAerialMap \cite{smith2015openaerialmap} & 456 \\
\textbf{Marigot} & 3.4 cm/px & 2018 & Drone & OpenAerialMap \cite{smith2015openaerialmap} & 387 \\
\textbf{Pichelin} & 3.4 cm/px &  2017 & Drone & GoCD & 149 \\
\textbf{Roseau} & 2.6 cm/px &  2017 &Drone & GoCD & 348 \\
\textbf{Salisbury} & 3-5 cm/px & 2018  & Drone & OpenAerialMap \cite{smith2015openaerialmap}  & 280 \\
\\[-1.5ex]
\hline \\[-1.5ex]
\begin{tabular}[c]{@{}l@{}}\textbf{Saint Lucia}\vspace{-0.1cm}\\ \begin{scriptsize}(nationwide)\end{scriptsize}\end{tabular} & 10.0 cm/px & 2022 & Aircraft & GoSL & 2,485 \\
\textbf{Castries} & 4.5 cm/px & 2019 & Drone & GPRH \cite{GPRH2019} & 1,084 \\
\textbf{Dennery} & 4.2 cm/px & 2019 & Drone & GPRH \cite{GPRH2019} & 742 \\
\textbf{Gros Islet} & 3.6 cm/px & 2019 & Drone & GPRH \cite{GPRH2019} & 864\\
\bottomrule
\end{tabular}
\label{tab:aerial-images}
\end{table}

\begin{table}[h]
\small
\centering
\caption{Test set (\%) results of different CNN architectures for roof type and roof material classification trained using (a) only Dominica data, (b) only Saint Lucia data, and (c) using a combination of Dominica and Saint Lucia. Models trained on the combined dataset are evaluated using the combined test sets of Dominica and Saint Lucia.}
\label{table:results1}
\begin{tabular}{@{}p{0.02cm}lcccc@{}}
\multicolumn{4}{@{}l}{\textbf{(a) Dominica}}\\
\toprule
& & \textbf{F1 score} & \textbf{Precision} & \textbf{Recall} &  \textbf{Accuracy} \\ \midrule
\multirow{4.25}*{\rotatebox{90}{\textbf{\begin{scriptsize} Roof Type\end{scriptsize}}}} 
& \textbf{VGG16} &  86.15 & 86.86 & 85.79  & 85.37 \\
& \textbf{ResNet50}   & 86.16 & 85.80 & 86.70 & 85.70 \\
& \textbf{Inceptionv3}  & 86.53 & 86.93 & 86.44 & 85.70\\
& \textbf{EfficientNet-B0}  & \textbf{87.08} & 87.06 & 87.37 & 86.41 \\
\midrule
\multirow{-1.15}*{\rotatebox{90}{\textbf{\begin{scriptsize} Roof Material\end{scriptsize}}}} 
& \textbf{VGG16} & 89.21 & 90.11 & 88.73 & 88.89 \\
& \textbf{ResNet50} & \textbf{89.50}  & 90.49 & 88.84 & 89.44 \\
& \textbf{Inceptionv3}  & 88.24 & 89.72 & 87.36  & 88.17 \\
& \textbf{EfficientNet-B0}  & 89.00 & 90.20 & 88.23 & 88.94 \\
\bottomrule
\end{tabular}
\medskip \\
\begin{tabular}{@{}p{0.02cm}lcccc@{}}
\multicolumn{4}{@{}l}{\textbf{(b) Saint Lucia}}\\
\toprule
&  & \textbf{F1 score} & \textbf{Precision} & \textbf{Recall}  & \textbf{Accuracy} \\ \midrule
\multirow{4.25}*{\rotatebox{90}{\textbf{\begin{scriptsize} Roof Type\end{scriptsize}}}}  
& \textbf{VGG16} & 88.02 & 91.29 & 85.36  & 93.10 \\
& \textbf{ResNet50} &  \textbf{89.49} & 94.11 & 85.97 & 94.28\\
& \textbf{Inceptionv3}  & 87.50 & 94.32 & 83.02 & 93.45 \\ 
& \textbf{EfficientNet-B0}  & 88.10 & 94.77 & 83.51  & 93.29\\
\midrule
\multirow{-1.15}*{\rotatebox{90}{\textbf{\begin{scriptsize} Roof Material\end{scriptsize}}}}  
& \textbf{VGG16} & 88.27 & 92.38 & 85.41  & 91.52 \\
& \textbf{ResNet50} & 91.55  & 92.99 & 90.28  & 93.20 \\
& \textbf{Inceptionv3}  & 91.42  & 93.28 & 89.91  & 93.15 \\ 
& \textbf{EfficientNet-B0}  & \textbf{91.71} & 93.81 & 90.00  & 93.79 \\
\bottomrule
\end{tabular}
\medskip \\
\begin{tabular}{@{}p{0.02cm}lcccc@{}}
\multicolumn{4}{@{}l}{\textbf{(c) Combined (Dominica + Saint Lucia)}}\\
\toprule
&  & \textbf{F1 score} & \textbf{Precision} & \textbf{Recall}  & \textbf{Accuracy} \\ \midrule
\multirow{4.25}*{\rotatebox{90}{\textbf{\begin{scriptsize} Roof Type\end{scriptsize}}}}  
& \textbf{VGG16} & 88.70 & 88.70 & 88.72 & 89.03  \\
& \textbf{ResNet50} & 88.71 & 88.63 & 88.87 & 89.24  \\
& \textbf{Inceptionv3}  & 89.38 & 89.29 & 89.62 & 89.60 \\
& \textbf{EfficientNet-B0}  & \textbf{90.03} & 90.07 & 90.25  & 90.45  \\
\midrule
\multirow{-1.15}*{\rotatebox{90}{\textbf{\begin{scriptsize} Roof Material\end{scriptsize}}}}  
& \textbf{VGG16} & 90.18 & 92.09 & 88.72 & 90.73  \\
& \textbf{ResNet50} & 89.82 & 91.76 & 88.36 & 90.80 \\
& \textbf{Inceptionv3} & \textbf{90.42} & 92.03 & 89.10 & 90.80 \\
& \textbf{EfficientNet-B0} & 90.09 & 92.37 & 88.45 & 91.09 \\
\bottomrule
\end{tabular}
\end{table}

\end{document}